\documentclass{article} 
\PassOptionsToPackage{table}{xcolor}
\usepackage{iclr2026_conference,times}

\usepackage{amsmath,amsfonts,bm}

\def\eqref#1{equation~\ref{#1}}

\def\1{\bm{1}}

\DeclareMathAlphabet{\mathsfit}{\encodingdefault}{\sfdefault}{m}{sl}
\SetMathAlphabet{\mathsfit}{bold}{\encodingdefault}{\sfdefault}{bx}{n}

\usepackage[most]{tcolorbox}
\usepackage{hyperref}
\usepackage{url}
\usepackage{xcolor}
\usepackage{graphicx}
\usepackage{float}
\usepackage{algorithm}
\usepackage{algorithmic}
\usepackage{multirow}
\usepackage{booktabs}

\definecolor{darkblue}{rgb}{0, 0, 0.5}
\definecolor{darkgreen}{rgb}{0.0, 0.5, 0.0}

\newcommand{\CATPO}{\textsc{CATPO }}
\newcommand{\TreeRPO}{\textsc{TreeRPO }}

\title{CATPO: Critique-Augmented Tree Policy Optimization}

\author{
Ayush Singh\thanks{Equal contribution.} \\
Vision and Language Group\\
Indian Institute of Technology Roorkee\\
Roorkee, Uttarakhand, India\\
\texttt{ayush\_s@mt.iitr.ac.in}
\And
Umang Goyal$^*$ \\
Vision and Language Group\\
Indian Institute of Technology Roorkee\\
Roorkee, Uttarakhand, India\\
\texttt{umang\_g@mt.iitr.ac.in}
\AND
Ankur Dahiya$^*$ \\
Vision and Language Group\\
Indian Institute of Technology Roorkee\\
Roorkee, Uttarakhand, India\\
\texttt{ankur\_d@mfs.iitr.ac.in}
}

\iclrfinalcopy 
\begin{document}

\maketitle
\lhead{} 

\begin{abstract}
Reinforcement learning with verifiable rewards (RLVR) has become a dominant paradigm for improving the reasoning capabilities of large language models (LLMs). Recent tree-based methods such as \TreeRPO extend flat trajectory sampling with tree-structured rollouts to obtain dense, step-level reward signals without a separate process reward model. However, not all trees are equally informative: trees where all leaves succeed, all leaves fail, or the policy already predicts the reward distribution contribute little to gradient updates, wasting compute. We introduce CATPO (Critique-Augmented Tree Policy Optimization), which diagnoses and addresses this waste at the tree level. \CATPO first scores each tree via a \emph{tree informativeness score} $F(T)$, combining leaf-outcome diversity with policy-reward decorrelation at zero extra compute. For dead-wrong trees where all branches fail, \CATPO applies \emph{critique-guided healing}: it locates the shallowest failure point, generates a natural-language critique, and grafts refined continuations to recover training signal. Finally, an \emph{informativeness-weighted loss} scales each tree's gradient contribution by its normalized score, concentrating parameter updates on the most informative trees while preserving overall gradient magnitude. Experiments on Qwen2.5-Math-1.5B trained with the MATH dataset show that \CATPO achieves 37.5\% macro accuracy across four benchmarks (AIME24, MATH-500, OlympiadBench, MinervaMath), improving over \TreeRPO by 1.9\% and GRPO by 4.8\%.
\end{abstract}

\begin{figure}[t]
  \centering
  \includegraphics[width=\textwidth]{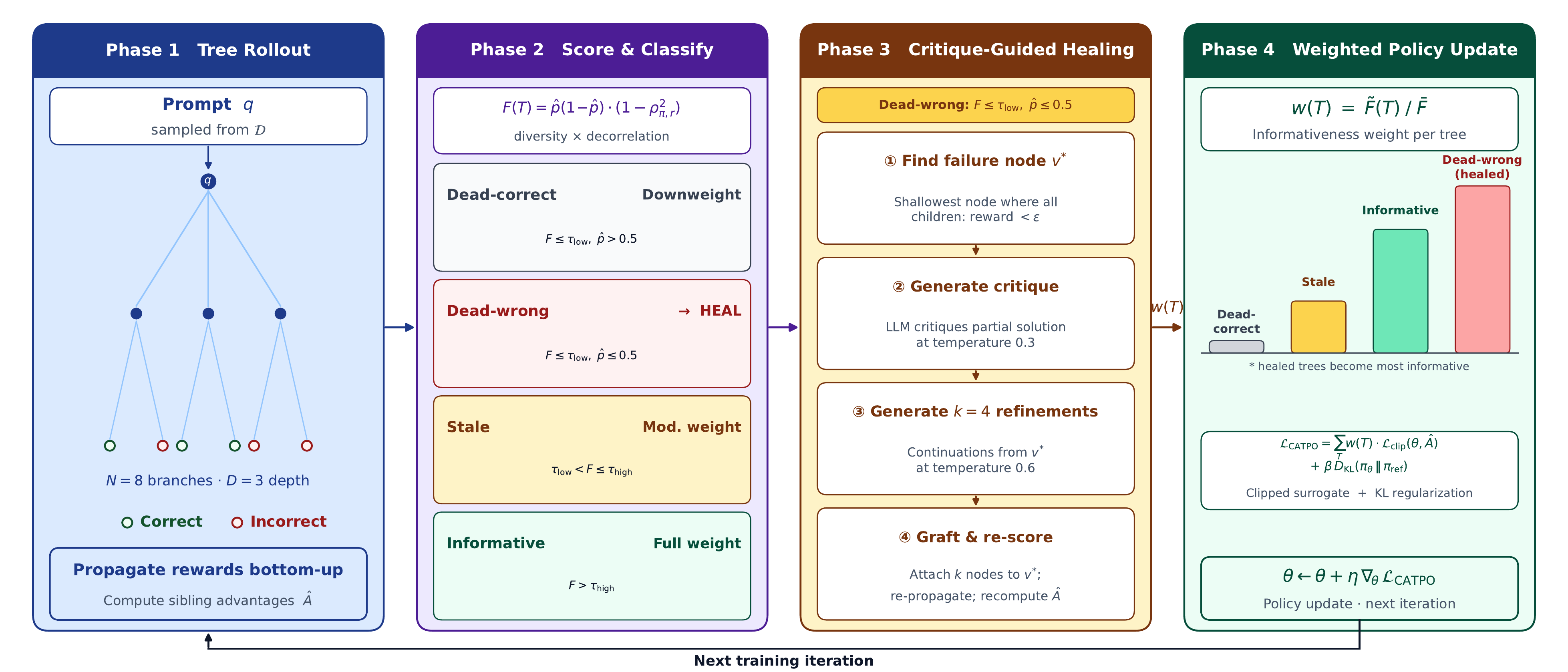}
  \caption{Overview of the \CATPO training pipeline. Each iteration proceeds in four phases: (1) sample an $N$-ary tree of partial solutions and propagate rewards; (2) score each tree with $F(T)$ and classify it into one of four regimes; (3) apply critique-guided healing to dead-wrong trees by finding the failure node, generating a critique, and grafting $k$ refined continuations; (4) scale advantages by the informativeness weight $w(T)$ and update the policy.}
  \label{fig:pipeline}
\end{figure}

\section{Introduction}
\label{sec:intro}

Recent breakthroughs in reinforcement learning have brought transformative progress to large language model (LLM) reasoning. Models such as DeepSeek-R1~\citep{guo2025deepseek}, and QwQ~\citep{qwq32b} achieve remarkable performance on complex reasoning tasks by training with Reinforcement Learning with Verifiable Rewards (RLVR), where the model explores reasoning paths toward correct, automatically checkable answers.

A central challenge in RLVR is credit assignment, trajectory-level rewards reveal whether a solution is right or wrong, but say little about which reasoning steps actually mattered. Two lines of work address this. Process reward models (PRMs)~\citep{verify_stepbystep, mathshepherd, omegaprm} train separate networks to score each step, but require costly step level annotation. Tree based methods avoid any auxiliary model, by expanding multiple continuations at each step, approaches like \TreeRPO~\citep{treerpo} and TreeRL~\citep{treerl} estimate step level advantages from leaf outcomes propagated bottom-up through the tree, yielding dense gradient signal at the cost of only extra sampling.

Yet tree-based methods treat all trees equally in the training loss. In practice, many sampled trees carry little or no useful signal:
\begin{itemize}
    \item \textbf{Dead-correct trees:} all leaves produce the right answer, the model already solves this problem, leaving no contrastive signal.
    \item \textbf{Dead-wrong trees:} every leaf fails, without a single positive branch, group-relative advantages collapse to zero.
    \item \textbf{Stale trees:} the policy's confidence already tracks the reward distribution, further training yields diminishing returns.
\end{itemize}

This waste is substantial. DAPO~\citep{dapo} and GRESO~\citep{greso} mitigate the analogous problem for flat rollouts by filtering or predicting uninformative prompts. But these operate at the prompt level, not the tree level: the same prompt can yield an informative tree or a dead one depending on the sample, and the internal structure of the tree - which branches the policy finds surprising - carries signal that prompt-level filtering discards.

We propose \textbf{\CATPO} (Critique-Augmented Tree Policy Optimization), which tackles this waste directly at the tree level through three complementary mechanisms:
\begin{enumerate}
    \item \textbf{Tree informativeness score.} We define $F(T) = \hat{p}(1{-}\hat{p}) \cdot (1 - \rho_{\pi,r}^2)$, combining leaf outcome diversity ($\hat{p}$: leaf pass rate) with policy-reward decorrelation ($\rho_{\pi,r}$: Pearson correlation between node log-probabilities and propagated rewards). Because every ingredient is already computed during tree sampling, $F(T)$ comes at zero additional cost.
    \item \textbf{Critique-guided healing.} For dead-wrong trees ($F \approx 0$), we locate the shallowest failing depth, prompt the policy model itself to critique the error, and graft $k$ critique conditioned continuations as new branches, recovering training signal from rollouts that would otherwise be wasted.
    \item \textbf{Informativeness-weighted loss.} Each tree's contribution to the policy gradient is scaled by its normalized score $w(T) = F(T) / \bar{F}$, modulating advantages and returns so that informative trees drive parameter updates while dead-correct and stale trees receive diminished influence.
\end{enumerate}

\section{Related Work}
\label{sec:related}

\subsection{Reinforcement Learning for LLM Reasoning}

Reinforcement learning from human feedback (RLHF)~\citep{rlhf} established the paradigm of optimizing LLM policies with reward signals from human preferences. PPO~\citep{ppo} has been the dominant algorithm, though its learned value function introduces instability and overhead. GRPO~\citep{deepseekmathgrpo} eliminated the value network entirely by using group-normalized rewards as advantage estimates, forming the backbone of breakthroughs like DeepSeek-R1~\citep{guo2025deepseek}. However, GRPO suffers from length and normalization biases~\citep{drgrpo}. Subsequent work has refined this paradigm: DAPO~\citep{dapo} introduced dynamic sampling and token level policy gradient losses, VAPO~\citep{vapo} advocated for value based credit assignment, and RLOO~\citep{rloo} and Open-Reasoner-Zero~\citep{openreasonerzero} showed that simpler REINFORCE style baselines can match or exceed GRPO.

\subsection{Tree-Based RL for Dense Reward Estimation}

Tree structured exploration has roots in inference time methods,  Tree-of-Thoughts~\citep{treeofthoughts}, RAP~\citep{rap}, and LATS~\citep{lats},  which explore multiple reasoning paths without updating model weights. ReST-MCTS*~\citep{restmcts} and rStar-Math~\citep{rstarmath} extended this idea to training time via MCTS for offline self-training. \TreeRPO~\citep{treerpo} was the first to integrate tree sampling into the on-policy RL loop, computing step level advantages via bottom-up reward propagation over sibling nodes. TreeRL~\citep{treerl} added entropy guided branching, while TreeGRPO~\citep{treegrpo} and Tree-OPO~\citep{treeopo} extended tree based advantages to agent tasks and off-policy settings. Crucially, all existing tree based methods treat every tree equally in the training objective. \CATPO is the first to \textit{diagnose} tree informativeness and \textit{intervene} on uninformative trees.

\subsection{Process Reward Models and Credit Assignment}

An alternative approach to step level credit assignment trains a separate neural network, a Process Reward Model (PRM), to score intermediate reasoning steps. Let's Verify Step by Step~\citep{verify_stepbystep} established that process supervision (per step correctness labels) outperforms outcome supervision for guiding the LLM policy, but requires expensive human annotation. Math-Shepherd~\citep{mathshepherd} and OmegaPRM~\citep{omegaprm} automated PRM data collection via Monte Carlo rollouts from intermediate states. PRIME~\citep{prime} computes token level implicit process rewards through a log ratio formulation without explicit PRM training. VinePPO~\citep{kazemnejad2025vineppo} replaces value networks with MC state value estimates for policy optimization, and SPO~\citep{spo} provides segment level credit assignment. These methods improve the quality of advantage estimates for the LLM policy gradient within existing rollout structures. \CATPO is complementary, operating at the tree level to filter and repair the rollouts themselves before they are used for policy updates.

\subsection{Sample Efficiency in LLM RL}

Wasted compute from uninformative rollouts is a recognized problem. DAPO's Dynamic Sampling~\citep{dapo} filters prompts where all responses succeed or fail. GRESO~\citep{greso} predicts uninformative prompts before rollout, achieving 2.4$\times$ speedup. CDAS~\citep{cdas} aligns problem difficulty to model competence. However, these methods operate at the prompt level, they decide whether to train on a problem at all, but cannot diagnose the internal structure of a tree rollout. \CATPO operates at the tree level, scoring and repairing the rollout structure itself.

\subsection{Self-Correction and Critique for LLMs}

Self-Refine~\citep{selfrefine} and Reflexion~\citep{reflexion} demonstrated iterative self-improvement through inference time critique, though \citet{huangselfcorrect} showed that intrinsic self-correction often degrades reasoning without external feedback. SCoRe~\citep{score} addressed this via multi-turn online RL, while GLoRe~\citep{glore}, RISE~\citep{rise}, and DRLC~\citep{drlc} trained refinement models on synthetic correction trajectories or used critiques as dense rewards.

\CATPO differs structurally, rather than correcting complete trajectories or training a separate refinement model, we apply critique within the tree at the specific point where reasoning first goes wrong, grafting corrected branches while preserving the shared prefix. This is more compute efficient than full regeneration and integrates naturally with the tree-based RL loop.

\section{Methodology}
\label{sec:method}

We present \CATPO in four parts: background on tree-based RL (Section~\ref{sec:background}), the tree informativeness score (Section~\ref{sec:fitness}), critique-guided healing (Section~\ref{sec:healing}), and the informativeness-weighted training objective (Section~\ref{sec:objective}). Figure~\ref{fig:pipeline} gives an overview of the full training pipeline.

\subsection{Background: Tree-Based RL with GRPO}
\label{sec:background}

\paragraph{GRPO.} Given a prompt $q$, GRPO~\citep{deepseekmathgrpo} samples a group of $G$ complete responses $\{o_i\}_{i=1}^G$ from the current LLM policy $\pi_{\theta_\text{old}}$. Each response receives a reward $R_i$ from a rule-based verification function (e.g., checking whether the final answer matches the ground truth). The advantage for response $i$ is computed by normalizing rewards within the group. The LLM policy parameters $\theta$ are then updated via a clipped surrogate objective with KL regularization against a reference policy $\pi_\text{ref}$. Crucially, no separate value network or reward model is trained - only the LLM policy is optimized.

\paragraph{TreeRPO.} Instead of sampling $G$ independent trajectories, \TreeRPO~\citep{treerpo} samples an $N$-ary tree of depth $D$ from the LLM policy. At each reasoning step, $N$ candidate continuations (each up to $L_\text{step}$ tokens) are generated, producing a tree where sibling nodes share a common prefix. Leaf nodes are scored by a verification function $\phi$, and rewards propagate bottom-up via mean aggregation:
\begin{equation}
    r_\text{node} = \mathbb{E}_{c \in \text{Children}(v)} [r_c].
    \label{eq:propagate}
\end{equation}

Advantages are computed over step-level sibling groups (children of the same parent). For a node with propagated reward $r_n$ and sibling rewards $\{r_s\}$, \TreeRPO uses a Bernoulli-variance normalization:
\begin{equation}
    \hat{A}_{n} = \frac{r_n - \mu}{\mu(1-\mu) + \epsilon}, \quad \mu = \text{mean}(\{r_s\}),
    \label{eq:treerpo_adv}
\end{equation}
where $\mu(1-\mu)$ replaces the empirical standard deviation to maintain consistency between binary and continuous reward normalization. This advantage is replicated across all tokens in the node's response. No separate value network or reward model is trained.

\paragraph{Limitation.} As discussed in Section~\ref{sec:intro}, both GRPO and \TreeRPO weight all sampled data equally. Zero-variance groups in GRPO become zero-gradient trees in \TreeRPO ,dead-correct, dead-wrong, and stale trees all waste compute. No existing mechanism diagnoses or repairs uninformative trees within the training loop; \CATPO fills this gap.

\subsection{Tree Informativeness Score}
\label{sec:fitness}

We propose a \emph{tree informativeness score} that quantifies how useful a tree $T$ is for improving the LLM policy. The score combines two signals available at zero extra cost after tree sampling:

\paragraph{Leaf Diversity Term.} Let $\hat{p}$ be the fraction of leaves in $T$ that receive a positive reward. The implementation uses Bernoulli variance:
\begin{equation}
    V_B(\hat{p}) = \hat{p}(1-\hat{p}).
    \label{eq:bern_var}
\end{equation}
This term is zero when all leaves agree (all correct or all incorrect), and reaches its maximum $0.25$ at $\hat{p}=0.5$. Intuitively, it measures the amount of contrastive signal available for group-relative advantages. 

\paragraph{Policy-Reward Decorrelation.} Let $\ell_n = \sum_t \log \pi_\theta(o_{n,t} \mid q, o_{n,<t})$ be the sum of token log-probabilities for non-root node $n$ under the current policy, and $r_n$ the propagated reward of node $n$. We compute:
\begin{equation}
    \rho_{\pi,r} = \text{Pearson}\big(\{\ell_n\}_{n \in T \setminus \text{root}},\ \{r_n\}_{n \in T \setminus \text{root}}\big).
    \label{eq:rho}
\end{equation}
We use $1-\rho_{\pi,r}^2$, which is the fraction of reward variation not linearly explained by policy confidence. When $|\rho_{\pi,r}| \approx 1$, the reward structure is largely explained and the tree is less informative; when $\rho_{\pi,r}$ is near $0$, learning potential is higher.

\paragraph{Informativeness Score.} Combining the leaf diversity term (Eq.~\ref{eq:bern_var}) with the decorrelation term (Eq.~\ref{eq:rho}), the tree informativeness score is:
\begin{equation}
    F(T) = \hat{p}(1-\hat{p}) \cdot (1 - \rho_{\pi,r}^2)
    \label{eq:fitness}
\end{equation}
with range $[0, 0.25]$. This multiplicative form ensures that both ingredients are necessary, a tree needs diverse outcomes and residual policy uncertainty to yield useful gradients. All quantities are already computed during tree sampling and log-probability evaluation, so $F(T)$ incurs zero additional forward passes.

\paragraph{Tree Classification.} Based on $F(T)$, we classify trees into four regimes using thresholds $\tau_\text{low}$ and $\tau_\text{high}$ (Table~\ref{tab:regimes}):

\begin{table}[h]
\centering
\small
\begin{tabular}{llll}
\toprule
\textbf{Regime} & \textbf{Condition} & \textbf{Interpretation} & \textbf{Action} \\
\midrule
Dead-correct & $F \le \tau_\text{low}$, $\hat{p}(1-\hat{p}) < 0.05$, $\hat{p} > 0.5$ & Near-unanimous success & Downweight \\
Dead-wrong & $F \le \tau_\text{low}$, $\hat{p}(1-\hat{p}) < 0.05$, $\hat{p} \le 0.5$ & Near-unanimous failure & Heal via critique \\
Stale & $\tau_\text{low} < F \le \tau_\text{high}$ & Moderate residual signal & Moderate weight \\
Informative & $F > \tau_\text{high}$ & High learning potential & Full weight \\
\bottomrule
\end{tabular}
\caption{Tree classification regimes based on the informativeness score $F(T)$. We use $\tau_\text{low}{=}0.025$ and $\tau_\text{high}{=}0.10$ for $F\in[0,0.25]$.}
\label{tab:regimes}
\end{table}

\subsection{Critique-Guided Healing}
\label{sec:healing}

Dead-wrong trees ($F \approx 0$, all leaves incorrect) represent problems where the model consistently fails. Rather than discarding them, \CATPO attempts to heal them by introducing corrected branches via natural-language critique.

\paragraph{Step 1: Identify the Failure Point.} We perform a breadth-first search from the root to find the shallowest depth $d^*$ where all children have propagated reward $r_c < \epsilon$ (with $\epsilon = 0.05$). This is the point where reasoning first goes entirely wrong.

\paragraph{Step 2: Generate a Critique.} The original problem text (from the root node) and the partial solution text (from the root to the failing node at depth $d^*$) are decoded. The policy model itself is prompted to identify the specific error:

\vspace{0.5em}
\begin{tcolorbox}[
  title=Critique Prompt,
  colback=gray!10,
  colframe=black,
  coltitle=white,
  colbacktitle=black,
  fonttitle=\bfseries,
  breakable,
  enhanced,
  boxed title style={
    sharp corners,
    boxrule=0pt
  }
]
\small
You are a mathematical reasoning critic. A student attempted the following problem but got it wrong.

Problem: \{Problem text\}

Student's partial solution: \{Partial solution\}

Task: Identify the specific mathematical or logical error. Be precise about which step is wrong and why.
\end{tcolorbox}
\vspace{0.5em}

The critique is generated at low temperature ($\tau = 0.3$) to produce a focused error diagnosis.

\paragraph{Step 3: Graft Refined Continuations.} The policy model is then prompted with the problem, the partial solution, and the generated critique, and asked to produce $k$ corrected continuations. These are scored by the verification function and added as new children of the failing node (at depth $d^*+1$), alongside the original failing children:

\vspace{0.5em}
\begin{tcolorbox}[
  title=Refinement Prompt,
  colback=gray!10,
  colframe=black,
  coltitle=white,
  colbacktitle=black,
  fonttitle=\bfseries,
  breakable,
  enhanced,
  boxed title style={
    sharp corners,
    boxrule=0pt
  }
]
\small
{Problem text}

A student's incorrect attempt: \{Partial solution\}

Critique of the error: \{Critique text\}

Provide a corrected solution continuing from where the error was found. Show your work step by step.
\end{tcolorbox}
\vspace{0.5em}

Each refinement is generated at the standard sampling temperature ($\tau = 0.6$), tokenized, truncated to $L_\text{step}$ tokens, and scored by the verification function to obtain a leaf reward.

\paragraph{Step 4: Re-score and Re-propagate.} The new leaves are scored by the verification function, rewards are re-propagated bottom-up via Eq.~\ref{eq:propagate}, and the informativeness score $F(T)$ is recomputed. If healing succeeds, some refined branches produce correct answers, converting a dead-wrong tree into one with positive reward diversity. We validate this empirically in Section~\ref{sec:experiments} (H2).

\subsection{CATPO Training Objective}
\label{sec:objective}

After healing, all nodes original and grafted are treated uniformly. Each tree's 
contribution to the policy gradient is weighted by its normalized informativeness:
\begin{equation}
    w(T) = \frac{F(\tilde{T})}{\frac{1}{|\mathcal{B}|}\sum_{T' \in \mathcal{B}} F(\tilde{T'})},
    \label{eq:fitness_weight}
\end{equation}
where $\tilde{T}$ denotes the tree after healing. This normalization preserves the overall
gradient magnitude while redistributing it toward informative trees. The weight $w(T)$ is treated as a detached constant during backpropagation; gradients do not flow through the informativeness computation. We adopt the clipped 
surrogate objective of \TreeRPO with this per-tree weight:
\begin{equation}
\begin{aligned}
    \mathcal{J}_\text{CATPO}(\theta) = \sum_{T \in \mathcal{B}} w(T) \sum_{n \in T} 
    \frac{1}{|o_n|} \sum_{t=1}^{|o_n|} \Big(
    &\min\big(r_{n,t}(\theta)\, \hat{A}_{n,t},\ 
    \text{clip}(r_{n,t}(\theta), 1-\varepsilon, 1+\varepsilon)\, \hat{A}_{n,t}\big) \\
    &- \beta\, D_\text{KL}(\pi_\theta \| \pi_\text{ref}) \Big),
\end{aligned}
\label{eq:catpo_loss}
\end{equation}
where $r_{n,t}(\theta)$ is the standard importance ratio and $\hat{A}_{n,t}$ is the 
Bernoulli-variance-normalized advantage (Eq.~\ref{eq:treerpo_adv}). Following \TreeRPO, 
only non-root nodes whose sibling reward range exceeds $\tau_\text{prune}=0.1$ participate. Grafted nodes from healing enter the same computation without modification, the informativeness
weight is the only difference from \TreeRPO. 

The complete algorithm is summarized in Appendix~\ref{app:algo}.
\section{Experiments}
\label{sec:experiments}

\subsection{Experimental Setup}

\paragraph{Datasets.} We follow the \TreeRPO experimental protocol. For training, we use the training split of the MATH dataset~\citep{hendrycksmath2021}, which contains 7.5k high-quality mathematical reasoning problems. For evaluation, we report results on four widely used benchmarks: MATH-500~\citep{verify_stepbystep} (500 representative problems from the MATH test split), OlympiadBench~\citep{olympiadbench}, MinervaMath~\citep{minerva}, and AIME24.

\paragraph{Base Models.} Our experiments use the Qwen2.5-Math series~\citep{qwen25math}. We report results on Qwen2.5-Math-1.5B.

\paragraph{Baselines.} We compare against:
\begin{itemize}
    \item \textbf{GRPO}~\citep{deepseekmathgrpo}: flat trajectory sampling with group-relative advantages.
    \item \textbf{\TreeRPO}~\citep{treerpo}: tree-structured sampling with step-level group-relative advantages (our direct base method).
\end{itemize}

\paragraph{Implementation Details.} We build on the rllm framework\footnote{\url{https://github.com/agentica-project/rllm}} derived from verl~\citep{verl}, with vLLM~\citep{vllm} for efficient inference. For GRPO: temperature 0.6, 8 rollouts per prompt, batch size 128, learning rate 1e-6, max response length 1152. For \TreeRPO and \CATPO: temperature 0.6, branching factor $N=8$, max depth $D=3$, max step length $L_\text{step}=384$, pruning threshold $\tau=0.1$. \CATPO-specific: $\tau_\text{low}=0.025$, $\tau_\text{high}=0.10$, dead-tree variance cutoff $\hat{p}(1-\hat{p})<0.05$, and $k=4$ refinements. KL coefficient $\beta=0.001$, entropy loss coefficient $\alpha=-0.001$. All experiments are conducted on a single A100-80GB GPU. Full hyperparameters and software details are provided in Appendix~\ref{app:implementation} (Tables~\ref{tab:hyperparams_opt}--\ref{tab:hyperparams_infra}).

\paragraph{Metrics.} We report \textbf{Pass@1 (Avg@8)} accuracy: for each test problem, 8 responses are sampled at temperature 0.6, and accuracy is averaged across samples. This differs from standard Pass@$k$, which measures whether \textit{any} of $k$ samples is correct; our metric reports the mean binary correctness across all 8 samples. We report individual benchmark scores and the \textbf{Macro Accuracy} (unweighted average across four benchmarks).

\begin{table}[t]
    \centering
    \renewcommand\arraystretch{1.3}
    \setlength{\tabcolsep}{5pt}
    \begin{tabular}{l|c|c|c|c|c}
        \toprule
        \textbf{Method} & \textbf{AIME24} & \textbf{MATH-500} & \textbf{OlympiadBench} & \textbf{Minerva} & \textbf{Macro Acc.} \\
        \midrule
        \rowcolor{gray!16} \multicolumn{6}{c}{\textit{Qwen2.5-Math-1.5B as the Base Model}} \\
        GRPO & 13.8 & 67.9 & 28.5 & 20.5 & 32.7 \\
        \TreeRPO & 16.8 {\scriptsize\color{darkgreen}$\uparrow$3.0} & 70.7 {\scriptsize\color{darkgreen}$\uparrow$2.8} & 30.9 {\scriptsize\color{darkgreen}$\uparrow$2.4} & 24.0 {\scriptsize\color{darkgreen}$\uparrow$3.5} & 35.6 {\scriptsize\color{darkgreen}$\uparrow$2.9} \\
        \textbf{\CATPO} & \textbf{20.3} {\scriptsize\color{darkgreen}$\uparrow$6.5} & \textbf{71.8} {\scriptsize\color{darkgreen}$\uparrow$3.9} & \textbf{33.1} {\scriptsize\color{darkgreen}$\uparrow$4.6} & \textbf{24.8} {\scriptsize\color{darkgreen}$\uparrow$4.3} & \textbf{37.5} {\scriptsize\color{darkgreen}$\uparrow$4.8} \\
        \bottomrule
    \end{tabular}
    \caption{Overall \textit{Pass@1} (\textit{Avg@8}) performance on four mathematical reasoning benchmarks.}
    \label{tab:main}
\end{table}

\subsection{Hypothesis Validation}

We validate the two core hypotheses that motivate our design: (H1) the informativeness score $F(T)$ is a meaningful proxy for how much a tree contributes to the policy gradient, and (H2) critique-guided healing can recover useful signal from dead-wrong trees.

\paragraph{H1: Informativeness Predicts Gradient Utility.}

We build trees offline for 100 problems from the MATH training set using Qwen2.5-Math-1.5B, compute $F(T)$ for each tree, and measure the per-tree gradient norm $\|g(T)\|^2$ via REINFORCE-style forward-backward passes. If $F(T)$ is a good proxy for training utility, it should correlate positively with $\|g(T)\|^2$.

\begin{table}[H]
\centering
\begin{tabular}{lcc}
\toprule
\textbf{Correlation Metric} & \textbf{Value} & \textbf{$p$-value} \\
\midrule
Pearson $r$ & 0.54 & 4.1e-05 \\
Spearman $\rho$ & 0.62 & 1.7e-06 \\
\bottomrule
\end{tabular}
\caption{Correlation between tree informativeness score $F(T)$ and per-tree gradient norm $\|g(T)\|^2$ on 50 trees from MATH training set.}
\label{tab:h1}
\end{table}

As shown in Table~\ref{tab:h1}, the tree informativeness score $F(T)$ exhibits a strong positive correlation with per-tree gradient norms, with Spearman $\rho = $ 0.62 ($p = $ 1.7e-06). The relationship is monotonic but not perfectly linear: trees with very low informativeness ($F < 0.025$) consistently produce near-zero gradient norms, confirming that dead-correct and dead-wrong trees are indeed uninformative for the policy update. In the mid-range ($0.025 < F < 0.10$), gradient norms increase steadily, while the most informative trees ($F > 0.10$) produce the largest gradients, though with greater variance, reflecting the fact that some informative trees contain harder problems where the gradient direction is useful but noisy. A small number of outliers appear at moderate $F$ with unusually high gradient norms, inspection reveals these correspond to problems at the boundary of the model's competence where a single correct branch among many failures creates a strong contrastive signal. Overall, H1 validates that $F(T)$ is a reliable, zero-cost proxy for gradient utility, justifying its use as an informativeness weight in the training loss.

\paragraph{H2: Critique Heals Dead-Wrong Trees.}

From the same tree set, we select all dead-wrong trees ($F \le 0.025$, $\hat{p}(1-\hat{p})<0.05$, $\hat{p} \le 0.5$) and apply critique-guided healing with $k=4$ refinements and measure the informativeness improvement percentage. 

Of the dead-wrong trees subjected to healing, 88.1\% show a positive improvement in $F$, indicating that the critique then refine pipeline successfully introduces at least one correct branch in the majority of cases. The remaining trees where healing fails ($\Delta F = 0$) correspond to problems that are fundamentally beyond the model's current capability: the critique correctly identifies the error, but the refined continuations repeat similar mistakes. This is consistent with prior findings that self-correction without external feedback is bounded by model capability~\citep{huangselfcorrect}. Notably, healing is most effective when the failure occurs at a shallow depth (depth 1), where the critique has access to a short, focused error trace rather than a long chain of compounding mistakes. Using the policy model as its own critic avoids the inference cost of a separate model while leveraging the policy's own understanding of its reasoning process. While this validates that critique-guided healing mechanistically restores gradient signal in dead-wrong trees, the downstream impact on policy accuracy is confirmed by the main results in Table~\ref{tab:main}.

\subsection{Main Results}

As shown in Table~\ref{tab:main}, \CATPO achieves a Macro Accuracy of 37.5\% on Qwen2.5-Math-1.5B, improving over \TreeRPO by 1.9\% and over GRPO by 4.8\%. The gains are consistent across all four benchmarks, but notably larger on the harder evaluation sets (AIME24 and OlympiadBench) than on MATH-500. This pattern is expected: harder problems produce a higher proportion of dead-wrong trees under the baseline, giving \CATPO more opportunities to recover training signal through both informativeness weighting and critique healing. On MATH-500, where the model already achieves reasonable accuracy under \TreeRPO, fewer trees fall into the dead-wrong regime, so the margin is naturally smaller. The substantial improvement on Minerva a benchmark outside the MATH training distribution suggests that informativeness-weighted training helps the model learn more transferable reasoning strategies rather than overfitting to the training domain.

\section{Conclusion}
\label{sec:conclusion}

We introduced \CATPO, a method that enhances tree-based reinforcement learning for training LLM reasoning policies through three complementary mechanisms, a zero-cost tree informativeness score that identifies which trees carry the most useful signal for the policy gradient, a critique-guided healing procedure that recovers training signal from dead-wrong trees, and an informativeness-weighted loss that concentrates parameter updates on informative trees. Our informativeness score $F(T)$ combines leaf-level diversity with policy-reward decorrelation to quantify each tree's learning potential, and we validated that it correlates with per-tree gradient norms and that critique healing significantly increases the informativeness of failing trees. By scaling the policy gradient loss with normalized tree informativeness, \CATPO redistributes gradient contributions toward the most useful rollouts while rescuing otherwise wasted compute through critique-guided healing.

\section*{Acknowledgments}
We thank E2E cloud networks for
providing the computational resources and GPUs
required for the project.

\bibliography{iclr2026_conference}

@article{guo2025deepseek,
  title={DeepSeek-R1: Incentivizing Reasoning Capability in LLMs via Reinforcement Learning},
  author={DeepSeek-AI and Guo, Daya and Yang, Dejian and Zhang, Haowei and Song, Junxiao and Wang, Peiyi and Zhu, Qihao and Xu, Runxin and Zhang, Ruoyu and Ma, Shirong and Bi, Xiao and Zhang, Xiaokang and Yu, Xingkai and Wu, Yu and Wu, Z. F. and Gou, Zhibin and Shao, Zhihong and Li, Zhuoshu and Gao, Ziyi and Liu, Aixin and Xue, Bing and Wang, Bingxuan and Wu, Bochao and Feng, Bei and Lu, Chengda and Zhao, Chenggang and Deng, Chengqi and Zhang, Chenyu and Ruan, Chong and Dai, Damai and Chen, Deli and Ji, Dongjie and Li, Erhang and Lin, Fangyun and Dai, Fucong and Luo, Fuli and Hao, Guangbo and Chen, Guanting and Li, Guowei and Zhang, H. and others},
  journal={arXiv preprint arXiv:2501.12948},
  year={2025},
  doi={10.1038/s41586-025-09422-z}
}

@misc{qwq32b,
    title = {QwQ-32B: Embracing the Power of Reinforcement Learning},
    author = {Qwen Team},
    month = {March},
    year = {2025},
    url = {https://qwenlm.github.io/blog/qwq-32b/}
}

@article{ppo,
  title={Proximal Policy Optimization Algorithms},
  author={Schulman, John and Wolski, Filip and Dhariwal, Prafulla and Radford, Alec and Klimov, Oleg},
  journal={arXiv preprint arXiv:1707.06347},
  year={2017}
}

@article{deepseekmathgrpo,
  title={DeepSeekMath: Pushing the Limits of Mathematical Reasoning in Open Language Models},
  author={Shao, Zhihong and Wang, Peiyi and Zhu, Qihao and Xu, Runxin and Song, Junxiao and Bi, Xiao and Zhang, Haowei and Zhang, Mingchuan and Li, Y. K. and Wu, Y. and Guo, Daya},
  journal={arXiv preprint arXiv:2402.03300},
  year={2024}
}

@misc{dapo,
  title={DAPO: An Open-Source LLM Reinforcement Learning System at Scale},
  author={Qiying Yu and Zheng Zhang and Ruofei Zhu and Yufeng Yuan and Xiaochen Zuo and Yu Yue and Weinan Dai and Tiantian Fan and Gaohong Liu and Lingjun Liu and Xin Liu and Haibin Lin and Zhiqi Lin and Bole Ma and Guangming Sheng and Yuxuan Tong and Chi Zhang and Mofan Zhang and Wang Zhang and Hang Zhu and Jinhua Zhu and Jiaze Chen and Jiangjie Chen and Chengyi Wang and Hongli Yu and Yuxuan Song and Xiangpeng Wei and Hao Zhou and Jingjing Liu and Wei-Ying Ma and Ya-Qin Zhang and Lin Yan and Mu Qiao and Yonghui Wu and Mingxuan Wang},
  year={2025},
  eprint={2503.14476},
  archivePrefix={arXiv},
  primaryClass={cs.LG},
  journal={https://arxiv.org/abs/2503.14476}
}

@misc{vapo,
  title={VAPO: Efficient and Reliable Reinforcement Learning for Advanced Reasoning Tasks},
  author={Yu Yue and Yufeng Yuan and Qiying Yu and Xiaochen Zuo and Ruofei Zhu and Wenyuan Xu and Jiaze Chen and Chengyi Wang and TianTian Fan and Zhengyin Du and Xiangpeng Wei and Xiangyu Yu and Gaohong Liu and Juncai Liu and Lingjun Liu and Haibin Lin and Zhiqi Lin and Bole Ma and Chi Zhang and Mofan Zhang and Wang Zhang and Hang Zhu and Ru Zhang and Xin Liu and Mingxuan Wang and Yonghui Wu and Lin Yan},
  year={2025},
  eprint={2504.05118},
  archivePrefix={arXiv},
  primaryClass={cs.AI},
  journal={https://arxiv.org/abs/2504.05118}
}

@article{rlhf,
  title={Training language models to follow instructions with human feedback},
  author={Ouyang, Long and Wu, Jeff and Jiang, Xu and Almeida, Diogo and Wainwright, Carroll L. and Mishkin, Pamela and Zhang, Chong and Agarwal, Sandhini and Slama, Katarina and Ray, Alex and Schulman, John and Hilton, Jacob and Kelton, Fraser and Miller, Luke and Simens, Maddie and Askell, Amanda and Welinder, Peter and Christiano, Paul and Leike, Jan and Lowe, Ryan},
  journal={Advances in Neural Information Processing Systems},
  volume={35},
  pages={27730--27744},
  year={2022},
  eprint={2203.02155},
  archivePrefix={arXiv},
  primaryClass={cs.CL}
}

@misc{drgrpo,
  title={Understanding R1-Zero-Like Training: A Critical Perspective},
  author={Zichen Liu and Changyu Chen and Wenjun Li and Penghui Qi and Tianyu Pang and Chao Du and Wee Sun Lee and Min Lin},
  year={2025},
  eprint={2503.20783},
  archivePrefix={arXiv},
  primaryClass={cs.LG},
  journal={https://arxiv.org/abs/2503.20783}
}

@misc{rloo,
  title={Back to Basics: Revisiting REINFORCE Style Optimization for Learning from Human Feedback in LLMs},
  author={Arash Ahmadian and Chris Cremer and Matthias Gall\'{e} and Marzieh Fadaee and Julia Kreutzer and Olivier Pietquin and Ahmet \"{U}st\"{u}n and Sara Hooker},
  year={2024},
  eprint={2402.14740},
  archivePrefix={arXiv},
  primaryClass={cs.LG},
  journal={https://arxiv.org/abs/2402.14740}
}

@misc{openreasonerzero,
  title={Open-Reasoner-Zero: An Open Source Approach to Scaling Up Reinforcement Learning on the Base Model},
  author={Jingcheng Hu and Yinmin Zhang and Qi Han and Daxin Jiang and Xiangyu Zhang and Heung-Yeung Shum},
  year={2025},
  eprint={2503.24290},
  archivePrefix={arXiv},
  primaryClass={cs.LG},
  journal={https://arxiv.org/abs/2503.24290}
}

@misc{treerpo,
  title={TreeRPO: Tree Relative Policy Optimization},
  author={Zhicheng Yang and Zhijiang Guo and Yinya Huang and Xiaodan Liang and Yiwei Wang and Jing Tang},
  year={2025},
  eprint={2506.05183},
  archivePrefix={arXiv},
  primaryClass={cs.LG},
  journal={https://arxiv.org/abs/2506.05183}
}

@misc{treerl,
  title={TreeRL: LLM Reinforcement Learning with On-Policy Tree Search},
  author={Zhenyu Hou and Ziniu Hu and Yujiang Li and Rui Lu and Jie Tang and Yuxiao Dong},
  year={2025},
  eprint={2506.11902},
  archivePrefix={arXiv},
  primaryClass={cs.LG},
  journal={https://arxiv.org/abs/2506.11902}
}

@misc{treegrpo,
  title={Tree Search for LLM Agent Reinforcement Learning},
  author={Yuxiang Ji and Ziyu Ma and Yong Wang and Guanhua Chen and Xiangxiang Chu and Liaoni Wu},
  year={2025},
  eprint={2509.21240},
  archivePrefix={arXiv},
  primaryClass={cs.LG},
  journal={https://arxiv.org/abs/2509.21240}
}

@misc{treeopo,
  title={Tree-OPO: Off-policy Monte Carlo Tree-Guided Advantage Optimization for Multistep Reasoning},
  author={Bingning Huang and Tu Nguyen and Matthieu Zimmer},
  year={2025},
  eprint={2509.09284},
  archivePrefix={arXiv},
  primaryClass={cs.AI},
  journal={https://arxiv.org/abs/2509.09284}
}

@inproceedings{treeofthoughts,
  title={Tree of Thoughts: Deliberate Problem Solving with Large Language Models},
  author={Yao, Shunyu and Yu, Dian and Zhao, Jeffrey and Shafran, Izhak and Griffiths, Thomas L. and Cao, Yuan and Narasimhan, Karthik},
  booktitle={Advances in Neural Information Processing Systems (NeurIPS)},
  year={2023}
}

@inproceedings{rap,
  title={Reasoning with Language Model is Planning with World Model},
  author={Hao, Shibo and Gu, Yi and Ma, Haodi and Hong, Joshua Jiahua and Wang, Zhen and Wang, Daisy Zhe and Hu, Zhiting},
  booktitle={Proceedings of the 2023 Conference on Empirical Methods in Natural Language Processing (EMNLP)},
  year={2023},
  eprint={2305.14992},
  archivePrefix={arXiv},
  primaryClass={cs.CL}
}

@article{lats,
  title={Language Agent Tree Search Unifies Reasoning Acting and Planning in Language Models},
  author={Zhou, Andy and Yan, Kai and Shlapentokh-Rothman, Michal and Wang, Haohan and Wang, Yu-Xiong},
  journal={arXiv preprint arXiv:2310.04406},
  year={2023},
  eprint={2310.04406},
  archivePrefix={arXiv},
  primaryClass={cs.AI}
}

@inproceedings{restmcts,
  title={{ReST-MCTS*}: {LLM} Self-Training via Process Reward Guided Tree Search},
  author={Zhang, Dan and Zhoubian, Sining and Hu, Ziniu and Yue, Yisong and Dong, Yuxiao and Tang, Jie},
  booktitle={Advances in Neural Information Processing Systems (NeurIPS)},
  year={2024},
  eprint={2406.03816},
  archivePrefix={arXiv},
  primaryClass={cs.CL}
}

@article{rstarmath,
  title={rStar-Math: Small {LLMs} Can Master Math Reasoning with Self-Evolved Deep Thinking},
  author={Guan, Xinyu and Zhang, Li Lyna and Liu, Yifei and Shang, Ning and Sun, Youran and Zhu, Yi and Yang, Fan and Yang, Mao},
  journal={arXiv preprint arXiv:2501.04519},
  year={2025},
  eprint={2501.04519},
  archivePrefix={arXiv},
  primaryClass={cs.AI}
}

@inproceedings{verify_stepbystep,
  title={Let's Verify Step by Step},
  author={Lightman, Hunter and Kosaraju, Vineet and Burda, Yura and Edwards, Harri and Baker, Bowen and Lee, Teddy and Leike, Jan and Schulman, John and Sutskever, Ilya and Cobbe, Karl},
  booktitle={International Conference on Learning Representations (ICLR)},
  year={2024}
}

@inproceedings{mathshepherd,
  title={Math-Shepherd: Verify and Reinforce {LLM}s Step-by-step without Human Annotations},
  author={Wang, Peiyi and Li, Lei and Shao, Zhihong and Xu, Runxin and Dai, Damai and Li, Yifei and Chen, Deli and Wu, Yu and Sui, Zhifang},
  booktitle={Proceedings of the 62nd Annual Meeting of the Association for Computational Linguistics (ACL)},
  year={2024},
  pages={9426--9439}
}

@article{omegaprm,
  title={Improve Mathematical Reasoning in Language Models by Automated Process Supervision},
  author={Luo, Liangchen and Liu, Yinxiao and Liu, Rosanne and Phatale, Samrat and Guo, Meiqi and Lara, Harsh and Li, Yunxuan and Shu, Lei and Zhu, Yun and Meng, Lei and Sun, Jiao and Rastogi, Abhinav},
  journal={arXiv preprint arXiv:2406.06592},
  year={2024},
  eprint={2406.06592},
  archivePrefix={arXiv},
  primaryClass={cs.CL}
}

@article{prime,
  title={Process Reinforcement through Implicit Rewards},
  author={Cui, Ganqu and Yuan, Lifan and Wang, Zefan and Wang, Hanbin and Zhang, Yuchen and Chen, Jiacheng and Li, Wendi and He, Bingxiang and Fan, Yuchen and Yu, Tianyu and Xu, Qixin and Chen, Weize and Yuan, Jiarui and Chen, Huayu and Zhang, Kaiyan and Lv, Xingtai and Wang, Shuo and Yao, Yuan and Han, Xu and Peng, Hao and Cheng, Yu and Liu, Zhiyuan and Sun, Maosong and Zhou, Bowen and Ding, Ning},
  journal={arXiv preprint arXiv:2502.01456},
  year={2025},
  eprint={2502.01456},
  archivePrefix={arXiv},
  primaryClass={cs.CL}
}

@inproceedings{kazemnejad2025vineppo,
  title={{VinePPO}: Refining Credit Assignment in {RL} Training of {LLMs}},
  author={Kazemnejad, Amirhossein and Aghajohari, Milad and Portelance, Eva and Sordoni, Alessandro and Reddy, Siva and Courville, Aaron and Le Roux, Nicolas},
  booktitle={Proceedings of the 42nd International Conference on Machine Learning (ICML)},
  year={2025},
  eprint={2410.01679},
  archivePrefix={arXiv},
  primaryClass={cs.LG}
}

@inproceedings{spo,
  title={Segment Policy Optimization: Effective Segment-Level Credit Assignment in {RL} for Large Language Models},
  author={Guo, Yiran and Xu, Lijie and Liu, Jie and Ye, Dan and Qiu, Shuang},
  booktitle={Advances in Neural Information Processing Systems (NeurIPS)},
  year={2025},
  eprint={2505.23564},
  archivePrefix={arXiv},
  primaryClass={cs.LG}
}

@article{greso,
  title={Act Only When It Pays: Efficient Reinforcement Learning for {LLM} Reasoning via Selective Rollouts},
  author={Zheng, Haizhong and Zhou, Yang and Bartoldson, Brian R. and Kailkhura, Bhavya and Lai, Fan and Zhao, Jiawei and Chen, Beidi},
  journal={arXiv preprint arXiv:2506.02177},
  year={2025},
  eprint={2506.02177},
  archivePrefix={arXiv},
  primaryClass={cs.LG}
}

@article{cdas,
  title={Rethinking the Sampling Criteria in Reinforcement Learning for {LLM} Reasoning: A Competence-Difficulty Alignment Perspective},
  author={Kong, Deyang and Guo, Qi and Xi, Xiangyu and Wang, Wei and Wang, Jingang and Cai, Xunliang and Zhang, Shikun and Ye, Wei},
  journal={arXiv preprint arXiv:2505.17652},
  year={2025},
  eprint={2505.17652},
  archivePrefix={arXiv},
  primaryClass={cs.CL}
}

@article{score,
  title={Training Language Models to Self-Correct via Reinforcement Learning},
  author={Kumar, Aviral and Zhuang, Vincent and Agarwal, Rishabh and Su, Yi and Co-Reyes, John D and Singh, Avi and Baumli, Kate and Iqbal, Shariq and Bishop, Colton and Roelofs, Rebecca and Zhang, Lei M and McKinney, Kay and Shrivastava, Disha and Paduraru, Cosmin and Tucker, George and Precup, Doina and Behbahani, Feryal and Faust, Aleksandra},
  journal={arXiv preprint arXiv:2409.12917},
  year={2024},
  eprint={2409.12917},
  archivePrefix={arXiv},
  primaryClass={cs.LG}
}

@article{selfrefine,
  title={Self-Refine: Iterative Refinement with Self-Feedback},
  author={Madaan, Aman and Tandon, Niket and Gupta, Prakhar and Hallinan, Skyler and Gao, Luyu and Wiegreffe, Sarah and Alon, Uri and Dziri, Nouha and Prabhumoye, Shrimai and Yang, Yiming and Gupta, Shashank and Majumder, Bodhisattwa Prasad and Hermann, Katherine and Welleck, Sean and Yazdanbakhsh, Amir and Clark, Peter},
  journal={Advances in Neural Information Processing Systems},
  volume={36},
  year={2023},
  eprint={2303.17651},
  archivePrefix={arXiv},
  primaryClass={cs.CL}
}

@article{reflexion,
  title={Reflexion: Language Agents with Verbal Reinforcement Learning},
  author={Shinn, Noah and Cassano, Federico and Berman, Edward and Gopinath, Ashwin and Narasimhan, Karthik and Yao, Shunyu},
  journal={Advances in Neural Information Processing Systems},
  volume={36},
  year={2023},
  eprint={2303.11366},
  archivePrefix={arXiv},
  primaryClass={cs.AI}
}

@article{huangselfcorrect,
  title={Large Language Models Cannot Self-Correct Reasoning Yet},
  author={Huang, Jie and Chen, Xinyun and Mishra, Swaroop and Zheng, Huaixiu Steven and Yu, Adams Wei and Song, Xinying and Zhou, Denny},
  journal={arXiv preprint arXiv:2310.01798},
  year={2023},
  eprint={2310.01798},
  archivePrefix={arXiv},
  primaryClass={cs.CL},
  note={Published at ICLR 2024}
}

@article{glore,
  title={GLoRe: When, Where, and How to Improve LLM Reasoning via Global and Local Refinements},
  author={Havrilla, Alex and Raparthy, Sharath and Nalmpantis, Christoforus and Dwivedi-Yu, Jane and Zhuravinskyi, Maksym and Hambro, Eric and Raileanu, Roberta},
  journal={arXiv preprint arXiv:2402.10963},
  year={2024},
  eprint={2402.10963},
  archivePrefix={arXiv},
  primaryClass={cs.CL}
}

@article{rise,
  title={Recursive Introspection: Teaching Language Model Agents How to Self-Improve},
  author={Qu, Yuxiao and Zhang, Tianjun and Garg, Naman and Kumar, Aviral},
  journal={arXiv preprint arXiv:2407.18219},
  year={2024},
  eprint={2407.18219},
  archivePrefix={arXiv},
  primaryClass={cs.LG}
}

@article{drlc,
  title={Beyond Sparse Rewards: Enhancing Reinforcement Learning with Language Model Critique in Text Generation},
  author={Cao, Meng and Shu, Lei and Yu, Lei and Zhu, Yun and Wichers, Nevan and Liu, Yinxiao and Meng, Lei},
  journal={arXiv preprint arXiv:2401.07382},
  year={2024},
  eprint={2401.07382},
  archivePrefix={arXiv},
  primaryClass={cs.CL}
}

@inproceedings{hendrycksmath2021,
  title={Measuring Mathematical Problem Solving With the MATH Dataset},
  author={Hendrycks, Dan and Burns, Collin and Kadavath, Saurav and Arora, Akul and Basart, Steven and Tang, Eric and Song, Dawn and Steinhardt, Jacob},
  booktitle={Advances in Neural Information Processing Systems (NeurIPS)},
  year={2021}
}

@inproceedings{olympiadbench,
  title={{O}lympiad{B}ench: A Challenging Benchmark for Promoting {AGI} with Olympiad-Level Bilingual Multimodal Scientific Problems},
  author={He, Chaoqun and Luo, Renjie and Bai, Yuzhuo and Hu, Shengding and Thai, Zhen Leng and Shen, Junhao and Hu, Jinyi and Han, Xu and Huang, Yujie and Zhang, Yuxiang and Liu, Jie and Qi, Lei and Liu, Zhiyuan and Sun, Maosong},
  booktitle={Proceedings of the 62nd Annual Meeting of the Association for Computational Linguistics (ACL)},
  year={2024}
}

@inproceedings{minerva,
  title={Solving Quantitative Reasoning Problems with Language Models},
  author={Lewkowycz, Aitor and Andreassen, Anders and Dohan, David and Dyer, Ethan and Michalewski, Henryk and Ramasesh, Vinay and Slone, Ambrose and Anil, Cem and Schlag, Imanol and Gutman-Solo, Theo and Wu, Yuhuai and Neyshabur, Behnam and Gur-Ari, Guy and Misra, Vedant},
  booktitle={Advances in Neural Information Processing Systems (NeurIPS)},
  year={2022}
}

@article{qwen25math,
  title={Qwen2.5-Math Technical Report: Toward Mathematical Expert Model via Self-Improvement},
  author={Yang, An and Zhang, Beichen and Hui, Binyuan and Gao, Bofei and Yu, Bowen and Li, Chengpeng and Liu, Dayiheng and Tu, Jianhong and Zhou, Jingren and Lin, Junyang and Lu, Keming and Xue, Mingfeng and Lin, Runji and Liu, Tianyu and Ren, Xingzhang and Zhang, Zhenru},
  journal={arXiv preprint arXiv:2409.12122},
  year={2024},
  eprint={2409.12122},
  archivePrefix={arXiv},
  primaryClass={cs.CL}
}

@inproceedings{verl,
  title={HybridFlow: A Flexible and Efficient RLHF Framework},
  author={Sheng, Guangming and Zhang, Chi and Ye, Zilingfeng and Wu, Xibin and Zhang, Wang and Zhang, Ru and Peng, Yanghua and Lin, Haibin and Wu, Chuan},
  booktitle={Proceedings of the Nineteenth European Conference on Computer Systems (EuroSys)},
  year={2025},
  doi={10.1145/3689031.3696075}
}

@inproceedings{vllm,
  title={Efficient Memory Management for Large Language Model Serving with PagedAttention},
  author={Kwon, Woosuk and Li, Zhuohan and Zhuang, Siyuan and Sheng, Ying and Zheng, Lianmin and Yu, Cody Hao and Gonzalez, Joseph E. and Zhang, Hao and Stoica, Ion},
  booktitle={Proceedings of the 29th Symposium on Operating Systems Principles (SOSP)},
  year={2023}
}
\bibliographystyle{iclr2026_conference}
\newpage

\appendix

\section{Algorithm}
\label{app:algo}
The complete algorithm is summarized in Algorithm~\ref{alg:catpo}.

\begin{algorithm}[H]
\caption{\CATPO: Critique-Augmented Tree Policy Optimization}
\label{alg:catpo}
\begin{algorithmic}[1]
\REQUIRE LLM policy $\pi_\theta$, reference policy $\pi_\text{ref}$, training data $\mathcal{D}$, branching factor $N$, max depth $D$, refinements $k$, thresholds $\tau_\text{low}, \tau_\text{high}$
\FOR{each training iteration}
    \STATE Sample batch of prompts $\{q_i\}$ from $\mathcal{D}$
    \STATE \textbf{Phase 1:} For each $q_i$, sample $N$-ary tree $T_i$ of depth $D$; score leaves; propagate rewards bottom-up; compute sibling-relative advantages (Eq.~\ref{eq:treerpo_adv}); prune low-variance sibling groups; compute $\log \pi_{\theta_\text{old}}$ for all non-pruned nodes
    \STATE \textbf{Phase 2:} Compute $F(T_i) = \hat{p}_i(1-\hat{p}_i) \cdot (1 - \rho_{\pi,r_i}^2)$ using the log-probs from Phase 1; classify each tree
    \STATE \textbf{Phase 3:} For each dead-wrong tree $T_i$ (where $F(T_i) \le \tau_\text{low}$, $\hat{p}_i(1-\hat{p}_i) < 0.05$, and $\hat{p}_i \le 0.5$):
    \STATE \quad Find shallowest node $v^*$ where all children have $r_c < \epsilon$
    \STATE \quad Decode problem text (from root) and partial solution (from root to $v^*$)
    \STATE \quad Generate critique at temperature $\tau{=}0.3$
    \STATE \quad Generate $k$ refined continuations at temperature $\tau{=}0.6$
    \STATE \quad Graft as children of $v^*$; score with verification function; re-propagate rewards; recompute advantages
    \STATE \textbf{Phase 4:} Compute informativeness weights $w(T_i) = F(\tilde{T}_i) / \overline{F}$; scale advantages by $w(T_i)$
    \STATE Update $\theta$ via $\nabla_\theta \mathcal{J}_\text{CATPO}(\theta)$ \hfill (Eq.~\ref{eq:catpo_loss})
\ENDFOR
\end{algorithmic}
\end{algorithm}

\section{Implementation Details}
\label{app:implementation}

We provide full implementation details for reproducibility. All experiments are built on the \texttt{rllm} framework\footnote{\url{https://github.com/agentica-project/rllm}} derived from \texttt{verl}~\citep{verl}, using vLLM~\citep{vllm} as the rollout engine with the XFORMERS attention backend. Training uses PyTorch 2.4 with FSDP for distributed training and gradient checkpointing enabled throughout.

\subsection{Model and Data}
\label{app:model_data}

We use Qwen2.5-Math-1.5B as the base model. The training data is the MATH dataset~\citep{hendrycksmath2021} formatted as chain-of-thought parquet files with \texttt{<think>} delimiters, which are ignored during reward computation (\texttt{ignore\_think=True}). Prompts are truncated to 512 tokens.

\subsection{Optimization Hyperparameters}
\label{app:opt_hyperparams}

Table~\ref{tab:hyperparams_opt} summarizes the optimization and batching hyperparameters.

\begin{table}[H]
\centering
\small
\renewcommand\arraystretch{1.15}
\caption{Optimization and batching hyperparameters.}
\begin{tabular}{lc}
\toprule
\textbf{Hyperparameter} & \textbf{Value} \\
\midrule
Learning rate & $1 \times 10^{-6}$ \\
LR warmup ratio & 0.0 \\
PPO epochs per iteration & 1 \\
PPO clip ratio $\varepsilon$ & 0.2 \\
PPO mini-batch size & 16 \\
Gradient clipping & 1.0 \\
KL loss type & low-var KL \\
KL loss coefficient $\beta$ & 0.001 \\
KL control coefficient & 0.001 \\
\midrule
Train batch size & 32 \\
Validation batch size & 128 \\
Max prompt length & 512 tokens \\
Max response length & 1152 tokens \\
Dynamic batch sizing & \checkmark \\
Max tokens per GPU & 12288 \\
\bottomrule
\end{tabular}
\label{tab:hyperparams_opt}
\end{table}

\subsection{Tree Structure and CATPO-Specific Hyperparameters}
\label{app:tree_hyperparams}

Table~\ref{tab:hyperparams_tree} details the hyperparameters governing tree rollouts (shared by TreeRPO and CATPO) and the CATPO-specific critique-healing parameters.

\begin{table}[H]
\centering
\small
\renewcommand\arraystretch{1.15}
\caption{Tree rollout and CATPO-specific hyperparameters.}
\begin{tabular}{lc}
\toprule
\textbf{Hyperparameter} & \textbf{Value} \\
\midrule
\multicolumn{2}{l}{\textit{Tree Structure (TreeRPO \& CATPO)}} \\
Branching factor $N$ & 8 \\
Step length $L_\text{step}$ & 384 tokens \\
Max depth $D = \lceil 1152/384 \rceil$ & 3 \\
Sampling temperature & 0.6 \\
Validation temperature & 0.6 \\
Validation rollouts $n_\text{val}$ & 8 \\
\midrule
\multicolumn{2}{l}{\textit{CATPO-Specific}} \\
Informativeness threshold $\tau_\text{low}$ & 0.025 \\
Informativeness threshold $\tau_\text{high}$ & 0.10 \\
Dead-tree variance cutoff & 0.05 \\
Healing failure threshold $\epsilon$ & 0.05 \\
Critique temperature & 0.3 \\
Critique max tokens & 256 \\
Refinement temperature & 0.6 \\
Number of refinements $k$ & 4 \\
\bottomrule
\end{tabular}
\label{tab:hyperparams_tree}
\end{table}

\subsection{Infrastructure}
\label{app:infra}

All experiments run on a single NVIDIA A100-80GB GPU. Table~\ref{tab:hyperparams_infra} lists infrastructure and training schedule settings.

\begin{table}[H]
\centering
\small
\renewcommand\arraystretch{1.15}
\caption{Infrastructure and training schedule.}
\begin{tabular}{lc}
\toprule
\textbf{Setting} & \textbf{Value} \\
\midrule
GPUs & 1$\times$A100-80GB \\
Tensor parallel size & 1 \\
FSDP parameter offload & \texttimes \\
FSDP optimizer offload & \checkmark \\
Reference model param offload & \checkmark \\
vLLM GPU memory utilization & 0.5 \\
Total epochs & 30 \\
Checkpoint frequency & every 20 steps \\
Evaluation frequency & every 20 steps \\
\bottomrule
\end{tabular}
\label{tab:hyperparams_infra}
\end{table}

\end{document}